\documentclass{article}
\usepackage{graphicx} 
\usepackage{multirow}
\usepackage{caption}
\usepackage{subcaption}
\usepackage{xcolor}
\usepackage{amsmath}
\usepackage{booktabs}

\usepackage[square,numbers]{natbib}
\usepackage[hidelinks]{hyperref}
\usepackage[final]{dlde_neurips_2022}

\title{Multiscale Neural Operators for Solving Time-Independent PDEs}
\author{Winfried Ripken\thanks{Equal contribution}\\
    Merantix Momentum\\
\And Lisa Coiffard\textsuperscript{*}\\
  Merantix Momentum\\
  \And Felix Pieper\textsuperscript{*}\\
  Merantix Momentum\\
  \And Sebastian Dziadzio\\
  Tübingen AI Center}
\date{September 2023}

\begin{document}

\maketitle
\begin{abstract}
Time-independent Partial Differential Equations (PDEs) on large meshes pose significant challenges for data-driven neural PDE solvers. We introduce a novel graph rewiring technique to tackle some of these challenges, such as aggregating information across scales and on irregular meshes. Our proposed approach bridges distant nodes, enhancing the global interaction capabilities of GNNs. Our experiments on three datasets reveal that GNN-based methods set new performance standards for time-independent PDEs on irregular meshes. Finally, we show that our graph rewiring strategy boosts the performance of baseline methods, achieving state-of-the-art results in one of the tasks.
\end{abstract}

\section{Introduction}

Partial Differential Equations (PDEs) describe a wide range of physical systems, particularly in thermodynamics, electromagnetics, or fluid dynamics \cite{cao2018liquid}. Traditional numerical solvers find functions that satisfy PDEs on a discretized solution space, i.e. using reference points on a mesh. Data-driven neural PDE solvers~\citep{bhattacharya2021model} reduce the computational overhead for obtaining solutions drastically, while generalizing to unseen boundary conditions and parameters \citep{li2020fourier, li2020multipole}.

\begin{figure}
    \centering
     \begin{subfigure}[t]{0.3\linewidth}
         \centering
         \includegraphics[width=\textwidth]{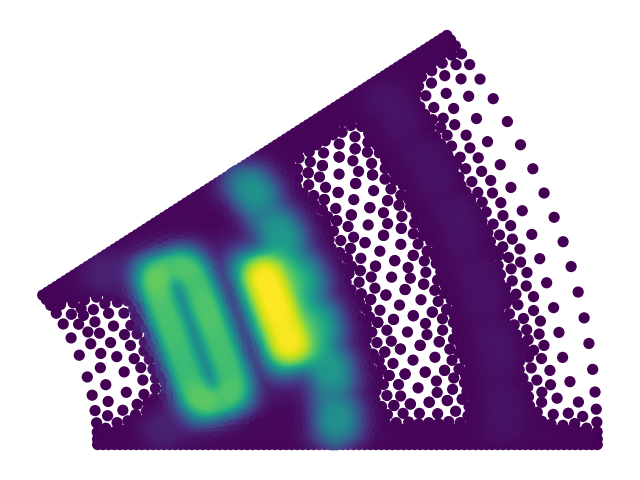}
         \caption{Node density via a Gaussian kernel-density estimate.}
     \end{subfigure}
     \hfill
     \begin{subfigure}[t]{0.3\textwidth}
         \centering
         \includegraphics[width=\textwidth]{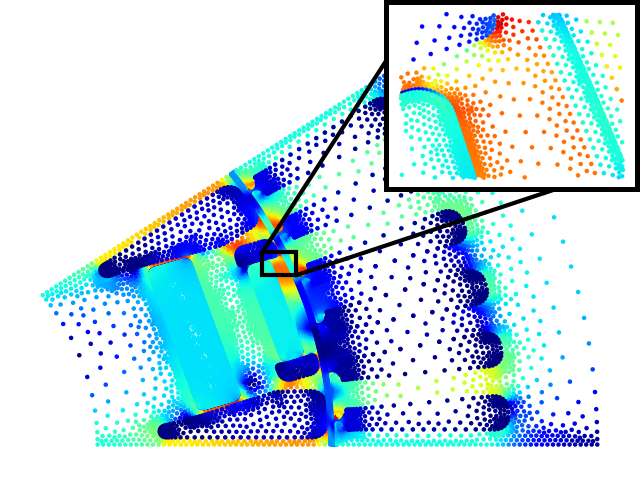}
         \caption{Discretized magnetic field (about 30K nodes).}
     \end{subfigure}
     \hfill
     \begin{subfigure}[t]{0.3\textwidth}
         \centering
         \includegraphics[width=\textwidth]{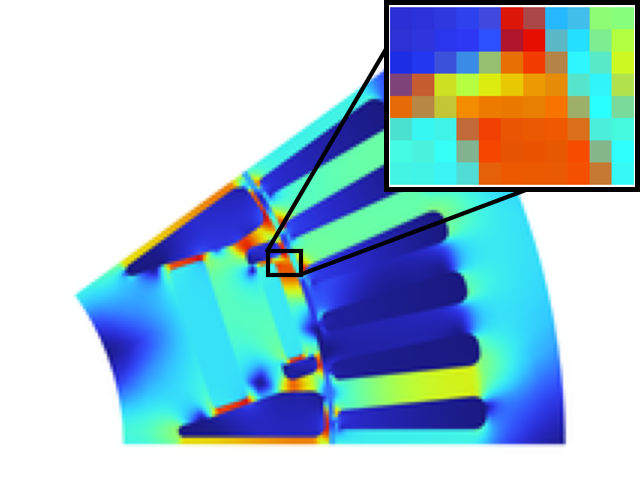}
         \caption{Magnetic field rasterized as pixel image of 30K pixels.}
     \end{subfigure}
    \caption{For changing mesh resolutions, rasterization either fails to capture local information or requires a vast number of pixels, increasing computational complexity. We show an example from the Motor dataset introduced by \citet{botache2023enhancing}.}
    \label{fig:three_graphs}
\end{figure}

Neural operators face unique challenges when solving time-independent PDEs. As opposed to time-dependent PDEs, for which solutions are rolled out incrementally over time \cite{brandstetter2021message}, the former predict solutions in a single step. Popular approaches to the time-independent setting include Fourier operators \citep{li2020fourier} and convolutional U-Nets \citep{gupta2022towards, ronneberger2015u}. However, these methods process information on a structured grid, which in many real-world scenarios fails to capture the different levels of precision that may be required over the spatial domain. We illustrate this in \autoref{fig:three_graphs}, where areas around the permanent magnets of the motor require higher mesh resolutions. 

In this work, we evaluate graph neural networks (GNNs) and transformers on existing PDE meshes. As the former capture only local interactions, we introduce a novel method for rewiring graphs to connect distant nodes (\autoref{sec:hierarchical_edges})\footnote[1]{Our source code is publicly available: \href{https://github.com/merantix-momentum/multiscale-pde-operators}{https://github.com/merantix-momentum/multiscale-pde-operators}}. Our contributions can be summarized as follows:

\begin{itemize}
    \item We benchmark existing GNN- and transformer-based methods on three challenging datasets: Darcy Flow~\citep{takamoto2022pdebench}, magnetic simulations of a synchronous electric motor~\citep{botache2023enhancing} and magnetostatics simulations of current-carrying conductors~\citep{lotzsch2022learning}.
    \item We propose a graph rewiring strategy that reaches state-of-the-art performance. 
    \item We propose a novel architecture that combines transformers with message passing, improving transformer-only results.
\end{itemize}

\section{Related Work}
\label{sec:related_work}

Neural networks for processing graph data have gained traction recently under the term \textit{geometric deep learning}~\citep{bronstein2017geometric}. \citet{gilmer2017neural} coin the term \textit{Message Passing Neural Networks} (MPNNs) to denote computing messages as a form of information exchange between graph nodes.

\paragraph{GNN-based models.} \citet{pfaff2020learning} introduce Mesh Graph Nets (\textbf{MGN}) to model physical systems by performing message passing on meshes. Their method consists of an encode-process-decode architecture: within the encoder, relative distances between nodes are added as edge attributes, to achieve spatial equivariance. 

\paragraph{Hierarchical GNN models.} Several works consider the limitations of GNNs in propagating information across long ranges within the graph. \citet{dwivedi2022long}. \citet{li2020multipole} are among the first to propose a hierarchical GNN operator for neural PDE solving. They create meshes of varying resolutions by randomly sub-sampling nodes, going from dense to coarse resolutions, and then back.

Similarly, \citet{gao2019graph} propose a U-Net-based architecture (g-UNet), with adaptive pooling and unpooling operations based on a learnable transformation of node features. A major limitation of this approach is the high probability of ending up with isolated nodes in down-sampled graphs.

\citet{cao2022bi} propose the Bi-stride Multi-Scale GNN (\textbf{BSMS}): they conduct a breadth-first search (BFS) on the graph at each stage and pool every second front of the BFS result. To provably leave no isolated partitions, they perform adjacency matrix enhancement, i.e. adding edges to the graph by using the second power of the adjacency matrix before every pooling step.

\paragraph{Graph rewiring methods.} Another family of methods investigates increasing graph connectivity between distant nodes. \citet{gutteridge2023drew} use a layer-dependent rewiring approach, such that edges are added over a maximum of $k$-hop neighbors at a distance $l+1$ for each layer $l$.

Transformer-based methods can be considered a \textit{global} form of rewiring, where every node is connected to others in the graph via attention. \textbf{Perceiver IO} \citep{jaegle2021perceiver} relies on cross-attention, similarly to other frameworks proposed for PDE solving \citep{li2023transformer}.


\section{Method: Creation of Hierarchical Tree Edges}
\label{sec:hierarchical_edges}

We hypothesize that processing information across different scales is required to effectively solve time-independent PDEs. 
This can be achieved in various ways, e.g. by combining full-scale attention with local processing, or in a hierarchical manner, as in g-UNets (\autoref{sec:related_work}). 
To empirically validate our claim, we design a novel graph rewiring strategy that augments any message passing method with the capability to process information across scales. 
Most notably, our procedure for adding \textit{hierarchical edges} to a graph depends only on node positions and is very lightweight, as it directly supplements edges to a graph instead of creating multiple graphs for different scales.

\begin{figure}
    \centering
    \includegraphics[width=0.6\linewidth]{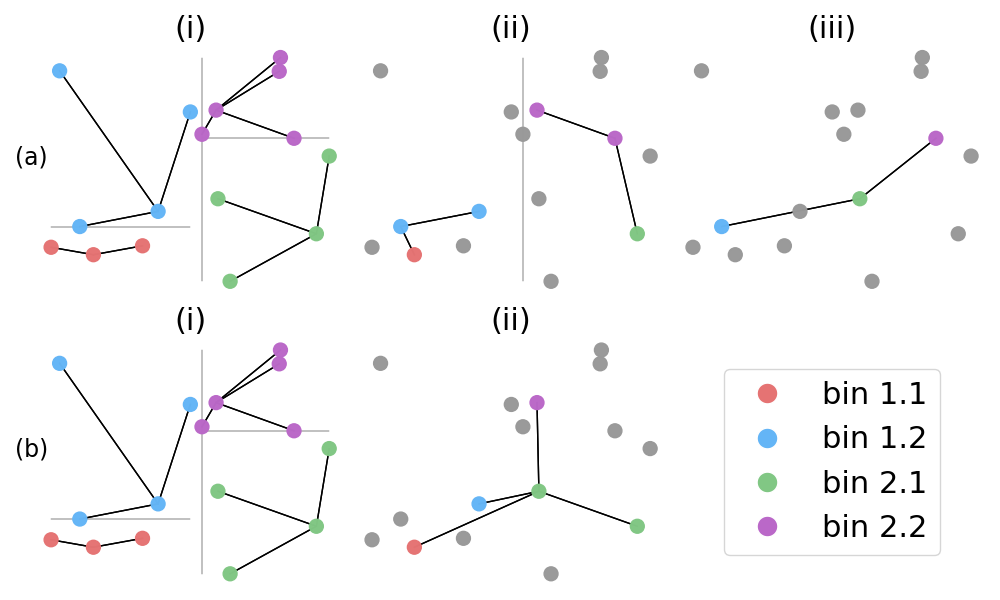}
    \caption{The construction of a binary tree for adding hierarchical edges in a graph, merging either \texttt{(a)} 2 bins, or \texttt{(b)} 4 bins at each level. Gray lines indicate split values; black lines indicate edges that are added to the computation graph. \texttt{(a,b)(i)} Creation of the binary tree: split along the $y$-axis to create two bins, then split these further along the $x$-axis. Connect all nodes in a bin to their respective center nodes. \texttt{(a)(ii)} Merge bins 1.1 and 1.2 into bin 1, bins 2.1 and 2.2 into bin 2, and add edges between the center nodes of the bins. \texttt{(a)(iii)} Merge remaining bins 1 and 2 and add edges between center nodes. \texttt{(b)(ii)} Merge all four bins and connect center nodes.}
    \label{fig:tree_simple}
\end{figure}

Creating \textit{hierarchical tree edges} (see \autoref{app:hierarchical_edges} for more details) involves recursively splitting graph nodes into bins according to their positions, over multiple levels (\autoref{fig:tree_simple}).
Our approach can be understood as a form of binary space partitioning~\citep{thibault1987set} for graphs.
Edges are added between the \texttt{center node} of each bin and that of its child bins. A \texttt{center node} is the node closest to the mean node position in a bin. 
The center nodes of the leaf bins are directly connected to all other nodes within the same bin (\autoref{fig:tree_simple}(a,b)(i)).
We introduce a separate hyperparameter that can skip levels in the binary tree to control the range of processing scales (\autoref{fig:tree_simple}), similarly to choosing the size of the receptive field in a U-Net.

Our rewiring technique can be combined with any GNN. While it is similar to BSMS in creating coarser graph representations to encourage the exchange of information over long distances, our method, contrary to BSMS, does not require pooling or up-sampling operations.


\section{Experiments}
\label{sec:experiments}

For our evaluation, we use three datasets: Darcy Flow, Motor, and Magnetostatics simulations (see \autoref{table:results_magn}, \ref{table:results_motor_darcy}). For Darcy Flow, we further include the results by \citet{takamoto2022pdebench} for reference. For the Magnetostatics experiments, we include results from \citet{li2023transformer} (a transformer-based method) and \citet{lotzsch2022learning} (a GNN method based on spectral convolutions).

\begin{table}[ht]
\centering
\begin{tabular}{lcc}
\toprule
Model    & Mag. Shape& Magn. Sup.\\
\midrule
ChebConv~\cite{defferrard2016convolutional} &  $6\times 10^{-06}$ & $1.72\times 10^{-04}$ \\
OFormer~\cite{li2023transformer} &  $1.6\times 10^{-05}$ & - \\
\midrule
MGN~\cite{pfaff2020learning} & $2.40\times 10^{-06}~(1.12\times 10^{-06})$ & $8.88\times 10^{-04}~(4.09\times 10^{-04})$\\
MGN + Tree (ours) & $\mathbf{1.31\times 10^{-06}~(1.46\times 10^{-07})}$ & $3.31\times 10^{-04}~(3.55\times 10^{-05})$\\
BSMS ($2 \times$MP)~\cite{gao2019graph} & $4.25\times 10^{-06}~(4.27\times 10^{-07})$ & $\mathbf{2.27\times 10^{-04}~(1.53\times 10^{-05})}$\\
Perceiver IO~\cite{jaegle2021perceiver} & $8.08\times 10^{-05}~(4.69\times 10^{-05})$ & $1.08\times 10^{-03}~(7.99\times 10^{-04})$\\
Perceiver IO + GNN & $3.52\times 10^{-06}~(1.55\times 10^{-06})$ & $2.73\times 10^{-04}~(1.26\times 10^{-04})$\\
\bottomrule
\end{tabular}
\vspace{0.5em}
\caption{Test MSE for baselines and custom methods (MGN + Tree, Perceiver IO + GNN) on the Magnetostatics dataset. We average over 3 runs, standard deviation in brackets. Results for ChebConv~\citep{defferrard2016convolutional} from~\citet{lotzsch2022learning} and OFormer from \citet{li2023transformer}. Best results in bold.}
\label{table:results_magn}
\end{table}

\begin{table}[ht]
\centering
\begin{tabular}{lcc}
\toprule
Model & Motor & Darcy Flow \\
\midrule
U-Net~\cite{ronneberger2015u} & $-$ & $\mathbf{6.4 \times 10^{-03}}$ \\
FNO~\cite{li2020fourier} & $-$ & $1.2 \times 10^{-02}$ \\
\midrule
MGN \cite{pfaff2020learning} & $9.36 \times 10^{-02}~(1.32\times 10^{-02})$ & $6.41 \times 10^{-02}~(5.23\times 10^{-04})$ \\
MGN + Tree (ours) & $7.55 \times 10^{-02}~(5.04\times 10^{-03})$ & $4.13 \times 10^{-02}~(1.47\times 10^{-02})$ \\
BSMS ($2 \times$MP)~\cite{gao2019graph} & $\mathbf{6.91 \times 10^{-02}~(2.78\times 10^{-03})}$ & $\mathbf{9.83 \times 10^{-03}~(8.18\times 10^{-04})}$ \\
Perceiver IO~\cite{jaegle2021perceiver} & $1.67 \times 10^{-01}~(2.18\times 10^{-02})$ & $1.41 \times 10^{-02}~(1.16\times 10^{-03})$ \\
Perceiver IO + GNN & $9.22 \times 10^{-02}~(8.01\times 10^{-03})$ & $1.31 \times 10^{-02}~(2.19\times 10^{-04})$ \\
\bottomrule
\end{tabular}
\vspace{0.5em}
\caption{Test RMSE of our baselines and custom methods (MGN + Tree, Perceiver IO + GNN) on the Motor and Darcy Flow datasets. All metrics are averaged over 3 runs with standard deviation in brackets. Results for U-Net and FNO from \citet{takamoto2022pdebench}. Best results are in bold; for Darcy Flow, we highlight both the best-performing method for regular grids (U-Net) and arbitrary grids (BSMS).}
\label{table:results_motor_darcy}
\end{table}

Our evaluations on Darcy Flow (\autoref{table:results_motor_darcy}) show that both GNN- and transformer-based methods achieve results that are in the same order of magnitude as convolution U-Nets and the Fourier Neural Operator (FNO). Note that, as all other datasets have non-uniform grids, we cannot apply convolution-based methods nor the FNO, underlining the necessity for methods that can process arbitrary meshes.
 
\paragraph{Hierarchical processing.}

We evaluate our rewiring approach by augmenting the input mesh of MGN with hierarchical tree edges (MGN+Tree) as described in \autoref{sec:hierarchical_edges}. 
Experiments show that adding our simple rewiring procedure improves upon MGN across all datasets. For the Magnetostatics Shape Generalization task, it achieves SOTA results, improving on the second-best performing method by a factor of 5 (\autoref{table:results_magn}).

For 3 out of 4 tasks, BSMS with 2 layers of message passing per scale achieves the best or second-best results, which supports our claim that hierarchical message passing is key to solving time-independent PDEs. The added advantage of introducing message passing layers at each level of the hierarchical operator in BSMS becomes significant for large graphs, such as the Darcy Flow dataset.

\paragraph{Comparison to full-scale attention.}
The results for Perceiver IO on the Darcy Flow dataset (\autoref{table:results_motor_darcy}) show that \textit{global} rewiring via cross-attention can perform almost on par with BSMS. For irregular grids with changing resolutions, both in the Motor and Magnetostatics datasets  (\autoref{table:results_motor_darcy} \& \ref{table:results_magn}), Perceiver IO falls behind in performance. 
We also augment Perceiver IO with message passing layers to capture local variations more accurately (Perceiver + GNN), resulting in substantially better performance. This suggests that the standard Perceiver IO may fail to capture local variations in high-resolution areas.

\paragraph{Further negative results.}
We found that the graph rewiring method proposed in DRew \cite{gutteridge2023drew} scales poorly to large graphs.
Additionally, we integrated our hierarchical tree edges (MGN+Tree) into a U-Net architecture (inspired by \cite{lino2022towards}), which did not yield improvements to our presented results.


\section{Summary \& Future Work}

In summary, all our experiments show that hierarchical information processing is key to solving time-independent PDEs on arbitrary meshes. Our results suggest that GNN-based and transformer-based methods can achieve similar performance to previously proposed techniques, which consider regular grids only. We believe that this finding is very relevant for practitioners, as irregular meshes are found frequently in real-world applications.



In the future, we hypothesize that our proposed Perceiver IO + GNN (\autoref{sec:experiments}) can achieve state-of-the-art results by improving the transformer's ability to distribute global information. Hyperparameter tuning per dataset could allow us to match the reported OFormer~\citep{li2023transformer} results. We also plan to investigate our proposed combination of graph U-Nets and hierarchical tree edges, using a binary tree that adapts to the changing resolution of the mesh, instead of rasterization on a fixed grid. Finally, as extensive aggregation of information over coarser scales leads to oversquashing~\citep{alon2020bottleneck}, we will study those effects in the context of hierarchical GNNs.

\section{Acknowledgements}
We would like to thank Martin Genzel and Maximilian Schambach for
insightful discussions, as well as Alma Lindborg and Sebastian Schulze for feedback on earlier stages of the manuscript.
We acknowledge funding by the Federal Ministry for Economic Affairs and Climate Action
(BMWK) within the project ”KITE: KI-basierte Topologieoptimierung elektrischer Maschinen” (\#19I21034B). The authors thank the International Max Planck Research School for Intelligent Systems (IMPRS-IS) for supporting Sebastian Dziadzio.

\bibliography{references}

\appendix
\section{Details for the Creation of Tree Edges}
\label{app:hierarchical_edges}

In this section, we describe our procedure for adding hierarchical edges to the graph in more detail.
Similarly to what is described in \citet{gladstone2023gnn} as edge augmentation, we add edges to the computation graph (via rewiring). Instead of adding edges randomly between pairs of nodes, we strive to add them in a structured hierarchical way, such that information can flow more efficiently across the graph.

First, we use the following procedure to split the graph nodes approximately in half, relying on the position information for all graph nodes. Let $n^i$ denote the position vector for node $i$. We identify the spatial dimension with the largest variance of node positions, i.e. we compute ${\arg\max}_{d\in\{x,y\}} \, \mathrm{Var}_i(n^i_d)\xrightarrow{}d$. We then split the nodes into two bins at the median of all node positions along that selected dimension: $s=\mathrm{median}_i(n^i_d)$ . We denote the node positions in the first bin as $\texttt{bin}_1 \xleftarrow{} \{n_{i} | n^i_d\geq s\}$ and for the second bin as $\texttt{bin}_2 \xleftarrow{} \{ n_i | n^i_d< s\}$. 
If we apply this procedure recursively $k$ times to create $k$ levels, we end up with a binary tree represented by all split values. Those split values are shown as the gray lines in \autoref{fig:tree_simple}. 

We now describe how we construct additional edges for the computation graph from this structure. We define the \texttt{center node} as the node in the graph, that is closest to the mean of nodes in a $\texttt{bin}$, i.e. $c(\texttt{bin})=\arg\min_{n_i\in \texttt{bin}} |n^i-\overline{n}|$ where $\overline{n}=\mathrm{mean}(\texttt{bin})$. Given $\texttt{bin}_1$ and $\texttt{bin}_2$ resulting from a split $s$, we add edges from node $c(\texttt{bin}_1\cup \texttt{bin}_2)$ to $c(\texttt{bin}_1)$ and $c(\texttt{bin}_2)$. 
Additionally, for the leaf level, all nodes of a bin are connected to the bins' center node. Doing this connects the whole graph hierarchically as displayed in \autoref{fig:tree_simple}. 

We also allow skipping over levels in the binary tree and merging $2^{n+1}$ bins in a single step. For example, skipping every two levels would mean connecting four bins (or nodes) to a single node (\autoref{fig:tree_simple}b) and so on.

For our experiments, we simply supplement all tree edges to the computation graph. Via the edge attributes (relative distances), the message passing operation is able to identify the hierarchical tree edges across larger distances.

\section{Dataset Details}
\paragraph{Darcy Flow.} We use the data made publicly available via PDEBench \citep{takamoto2022pdebench} for $\beta=1$. The task is to predict flow through a porous medium given the viscosity of a fluid. 

The Darcy flow dataset contains only uniform grids. It is challenging, because of a high grid resolution (about 16K graph nodes per sample) and therefore requires long-range message passing. 

\paragraph{Motor Simulations.} We use the data made publicly available by \citet{botache2023enhancing} for $\text{motor\_pos}=1$. Given motor topologies i.e. the arrangement of components (magnets, air cavities, coils, iron parts), the task is to predict the magnetic field. Even though the magnetic field inside the synchronous electric motor can change, it only depends on the applied current and the rotating angle, which we both keep constant in our case. We use Delaunay triangulation to obtain the meshes for our GNN-based methods. 

The motor dataset has a very high resolution (ca. 30K graph nodes per sample) and highly irregular meshes and therefore is particularly challenging. 

\paragraph{Magnetostatics Simulations.} We use the data made publicly available by \citet{lotzsch2022learning}. The task is to predict the magnetic vector potential from a distribution of electric currents. We omit the prediction of the magnetic field. The magnetostatics PDE used is a 2D Poisson equation with Dirichlet boundary conditions.

The magnetostatics dataset contains irregular meshes of 5 different shapes. It is challenging because it explicitly requires the network to generalize (the two test sets are distinct from the training set).

\section{Hyperparameters + Training Details}
For all methods, we use a node-wise MLP encoder and decoder before and after the operator, similar to \citet{pfaff2020learning}. For all experiments, the inputs are normalized with zero mean and unit standard deviation. All latent dimensions (node or edge embeddings, hidden layers of MLPs) are set to 64 if not specified otherwise.

We use the Adam optimizer \citep{kingma2014adam} with default parameters. For training, we calculate the mean squared error loss over all predicted quantities and use a virtual batch size of 16 for all experiments. The training is conducted on an NVIDIA T4 with 16GB VRAM in a Google Kubernetes cluster. All experiments are repeated 3 times over 200 epochs each. We save checkpoints every 3 epochs and perform testing on the checkpoint with the lowest validation loss (MSE). We use cosine annealing \citep{loshchilov2016sgdr} with $\eta_{min}=0$ as the learning rate schedule.

\paragraph{MGN \citep{pfaff2020learning}:} For Darcy Flow, we use the relative distances between nodes as edge attributes. As the problem is symmetric, we do not pass any absolute node positions, but rather the distance to the closest border in $x$- and $y$-direction. For the Motor data, we use relative distances between nodes as edge attributes and pass absolute positions as node attributes. We use 15 layers of message passing like in the original paper \citep{pfaff2020learning} and a latent dimension of 64 everywhere. 

\paragraph{MGN + Tree (ours):}
As edge attributes are initialized as relative distances, the hierarchical edges we add in \autoref{sec:hierarchical_edges} can be differentiated from the rest. For both the Motor dataset and Darcy Flow, we split 10 times and connect 8 bins to one (skipping 2 levels for constructing the added edges). For magnetostatics experiments, we split 4 times and connect all levels directly (2 bins are merged into one).

\paragraph{BSMS \citep{cao2022bi}:} For BSMS, we use the relative distances between nodes as edge attributes, as proposed in the original paper \citep{cao2022bi}. Similar to before, we pass the distance to the border for Darcy Flow and absolute positions for the Motor dataset as node attributes. We downsample the graph 5 times using breadth-first search and use one or two layers of message passing for each scale. 


\paragraph{Perceiver IO \citep{jaegle2021perceiver}:} We always pass absolute positions as node attributes and, for Darcy Flow, additionally the distance to the border. If we have to batch graphs of different sizes together, we zero-pad the node attributes. We re-use the input vector as the query vector for the last cross-attention layer, which has only been transformed by the very first node-wise MLP encoder. We use 128 latent tokens. We also add a skip connection between the MLP encoder and decoder, essentially bridging the Perceiver IO model together and enabling local information to flow through. This greatly enhances the performance of Perceiver IO, especially for the Motor dataset.

\paragraph{Perceiver IO + GNN:} We add 5 layers of message passing both before and after the Perceiver IO model. The output of the first message passing block is added to the output of the Perceiver IO model before feeding it to the second message passing block, adding a skip connection between the two blocks.

\section{Additional Results}
\begin{figure}
    \centering
    \includegraphics[width=\linewidth]{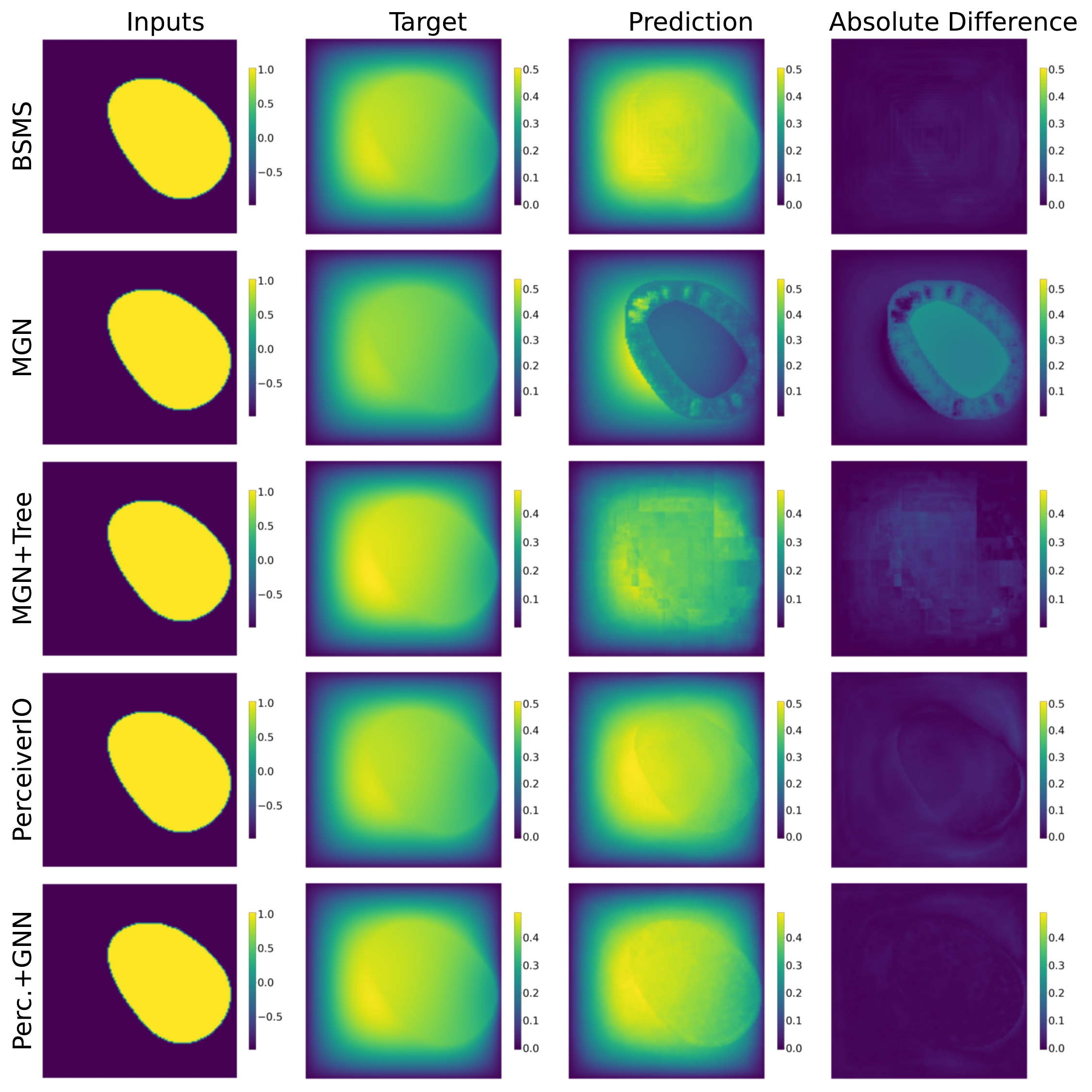}
    \caption{Example predictions for the Darcy Flow dataset \cite{takamoto2022pdebench}}
    \label{fig:darcy-viz}
\end{figure}

\begin{figure}
    \centering
    \includegraphics[width=\linewidth]{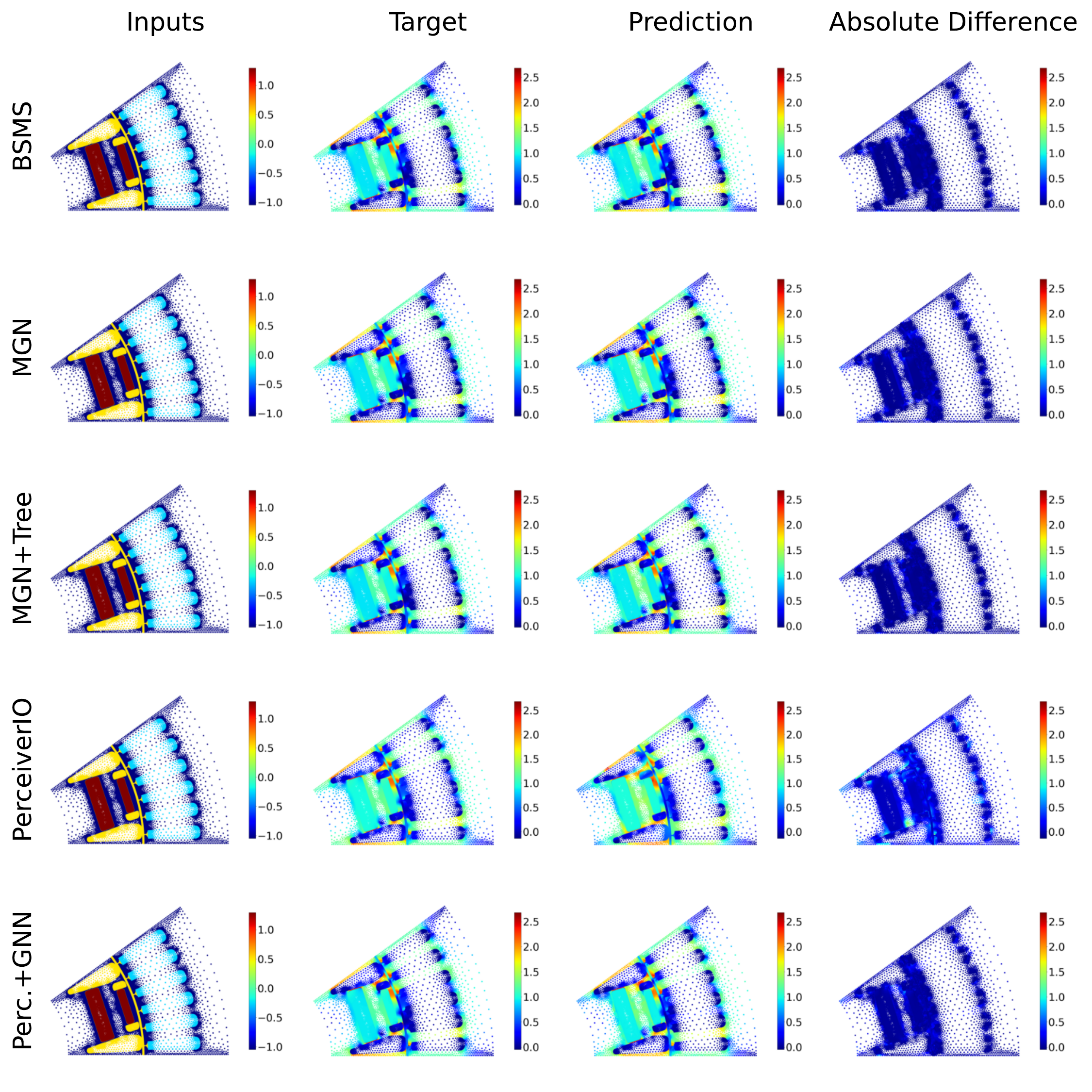}
    \caption{Example predictions for the electric Motor dataset \citep{botache2023enhancing}.}
    \label{fig:motor-viz}
\end{figure}

\begin{figure}
    \centering
    \includegraphics[width=\linewidth]{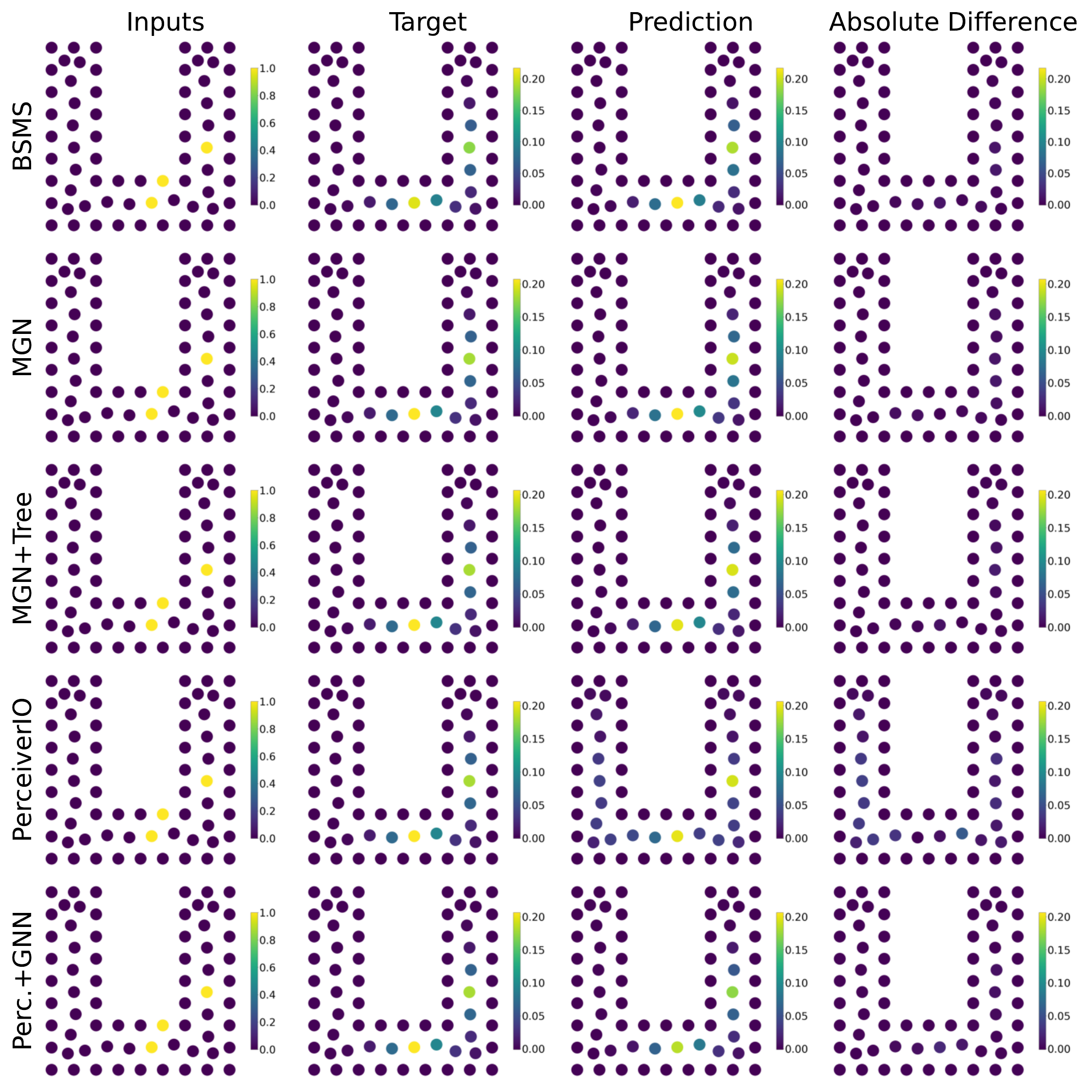}
    \caption{Example predictions for magnetostatics data on the shape generalization test set \cite{lotzsch2022learning}.}
    \label{fig:magn-viz-i}
\end{figure}

\begin{figure}
    \centering
    \includegraphics[width=\linewidth]{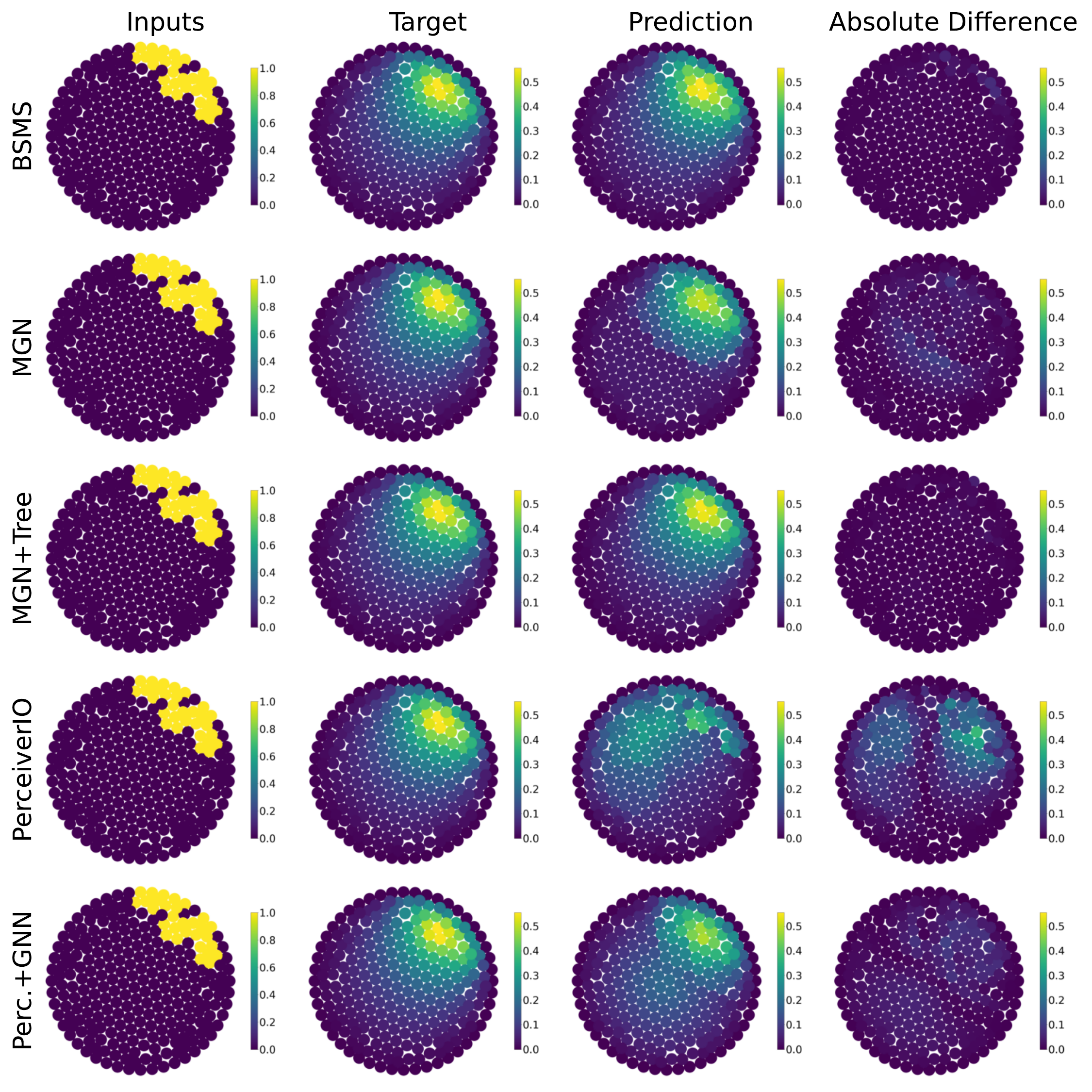}
    \caption{Example predictions for magnetostatics data on the superpositions test set \cite{lotzsch2022learning}.}
    \label{fig:magn-viz-ii}
\end{figure}

In \autoref{table:train_time}, we record the training time in hours for all of our methods. In \autoref{table:ablation_bsms} we compare the performance of BSMS with 2 layers against only one layer of message passing. We observe an improvement for most of the datasets and therefore report our main results using 2 layers of message passing. As improvements are relatively minor, this parameter does not seem to be crucial for the success of BSMS. In \autoref{fig:darcy-viz},  \autoref{fig:motor-viz}, \autoref{fig:magn-viz-i} and \autoref{fig:magn-viz-ii}, we visualize exemplary predictions for all methods and datasets.

\begin{table}[h]
\centering
\begin{tabular}{llll}
\toprule
Model    & Motor& Darcy Flow& Magnetostatics\\
\midrule
MGN & $18.95$ & $141.00$ & $2.65$\\
MGN + Tree & $25.91$ & $170.16$ & $3.04$\\
BSMS & $22.92$ & $161.81$ & $\mathbf{2.41}$\\
Perceiver IO & $\mathbf{3.10}$ & $\mathbf{17.46}$ & $5.25$\\
Perceiver IO + GNN & $15.46$ & $104.18$ & $6.27$\\
\bottomrule
\end{tabular}
\vspace{0.5em}
\caption{Average training time in hours, for our baselines and custom methods (MGN + Tree edges, Perceiver IO + GNN). All metrics are averaged over 3 runs on an Nvidia Tesla T4 GPU. For both magnetostatics tasks, we train only once, as only the test set changes. Perceiver IO is faster than GNN-based methods for datasets with large graphs, as we do not adapt the hyperparameters (e.g. number of latents).}
\label{table:train_time}
\end{table}

\begin{table}[ht]
\centering
\begin{tabular}{lllll}
\toprule
Model    & Motor& Darcy Flow& Mag. Shape& Magn. Sup.\\
\midrule
BSMS (2xMP) & $\mathbf{6.91\times 10^{-02}}$ & $9.83\times 10^{-03}$ & $\mathbf{1.84\times 10^{-03}}$ & $\mathbf{1.08\times 10^{-02}}$\\
BSMS (1xMP) & $7.29\times 10^{-02}$ & $\mathbf{9.48\times 10^{-03}}$ & $2.38\times 10^{-03}$ & $1.24\times 10^{-02}$\\
\bottomrule
\end{tabular}
\vspace{0.5em}
\caption{Ablation study of BSMS' number of message passing layers per scale. We report the mean RMSE scores averaged over 3 runs each.}
\label{table:ablation_bsms}
\end{table}

\end{document}